\documentclass{sig-alternate-05-2015}
\usepackage{graphicx}
\usepackage{amsmath}
\usepackage{amsfonts}
\usepackage{subcaption}
\usepackage{dblfloatfix} 
\usepackage{color}

\newcommand{\argmin}[1]{\underset{#1}{\operatorname{arg}\,\operatorname{min}}\;} 
\newcommand{\argmax}[1]{\underset{#1}{\operatorname{arg}\,\operatorname{max}}\;} 

\title{A Light-powered, ``Always-On'', Smart Camera with Compressed Domain Gesture Detection }
\advance\dimen0 by -12.75pc\relax
\author{
Anvesha A, Shaojie Xu, Ningyuan Cao, Justin Romberg and Arijit Raychowdhury \\
       \affaddr{School of Electrical and Computer Engineering, Georgia Institute of Technology}\\
       \email{$\{$aamaravati3, kyle.xu,ningyuan.cao$\}$@gatech.edu}\\
			\email{justin@ece.gatech.edu, arijit.raychowdhury@ece.gatech.edu}
}

\begin{document}

\CopyrightYear{2016} 
\setcopyright{acmcopyright}
\conferenceinfo{ISLPED '16,}{August 08-10, 2016, San Francisco Airport, CA, USA}
\isbn{978-1-4503-4185-1/16/08}\acmPrice{\$15.00}
\doi{http://dx.doi.org/10.1145/2934583.2934594}

\maketitle
\begin{abstract}
In this paper we propose an energy-efficient camera-based gesture recognition system powered by light energy for ``always on'' applications. Low energy consumption is achieved by directly extracting gesture features from the compressed measurements, which are the block averages and the linear combinations of the image sensor's pixel values. The gestures are recognized using a nearest-neighbour (NN) classifier followed by Dynamic Time Warping (DTW). The system has been implemented on an Analog Devices Black Fin ULP vision processor and powered by PV cells whose output is regulated by TI's DC-DC buck converter with Maximum Power Point Tracking (MPPT). Measured data reveals that with only $400$ compressed measurements ($768\times$ compression ratio) per frame, the system is able to recognize key wake-up gestures with greater than $80\%$ accuracy and only $95mJ$ of energy per frame. Owing to its fully self-powered operation, the proposed system can find wide applications in ``always-on'' vision systems such as in surveillance, robotics and consumer electronics with touch-less operation.
\end{abstract}

%
%

%
%

%
%



\section{Introduction}
As the ``Internet of Smart Things'' continues to have tremendous societal impacts, human-machine interfaces are also evolving in the modalities, accuracies and improved energy-efficiencies. Beyond the traditional keyboard and mice, such smart devices enable advanced user interfaces, like voice command and control, camera and GPS based sensors and interfaces, as well as touch screens and displays. However, these interfaces are mostly active, in the sense that they require significant power to receive user inputs and subsequently process them. Hence, in an ``always-on'' environment, where these user interfaces need to be perpetually ``on'', the design of the sensor front-ends and their power management present significant challenges. The power cost of continuously capturing and analyzing videos is so high that most systems require physical input from the user before accepting commands.  To address this issue, a ``wake up'' camera front-end allows a sensor node to continuously acquire videos and monitor for a trigger that will wake up the back-end when necessary, thus enabling exciting new usage models. A promising ``wake up'' modality in ``always on'' cameras is hand gestures, which is presented in this paper.
Traditional gesture recognition systems are power inefficient and run on batteries or even AC power supplies ~\cite{gesturefpga,leecpu}. However, with rapid advances in energy harvesting, it is enticing to think about a camera front-end which is powered by photo-voltaic cells (PV), thus paving the way for light-powered, smart, ``always on'' cameras. 

\begin{figure}[!b]
\centering
\includegraphics[scale=0.31]{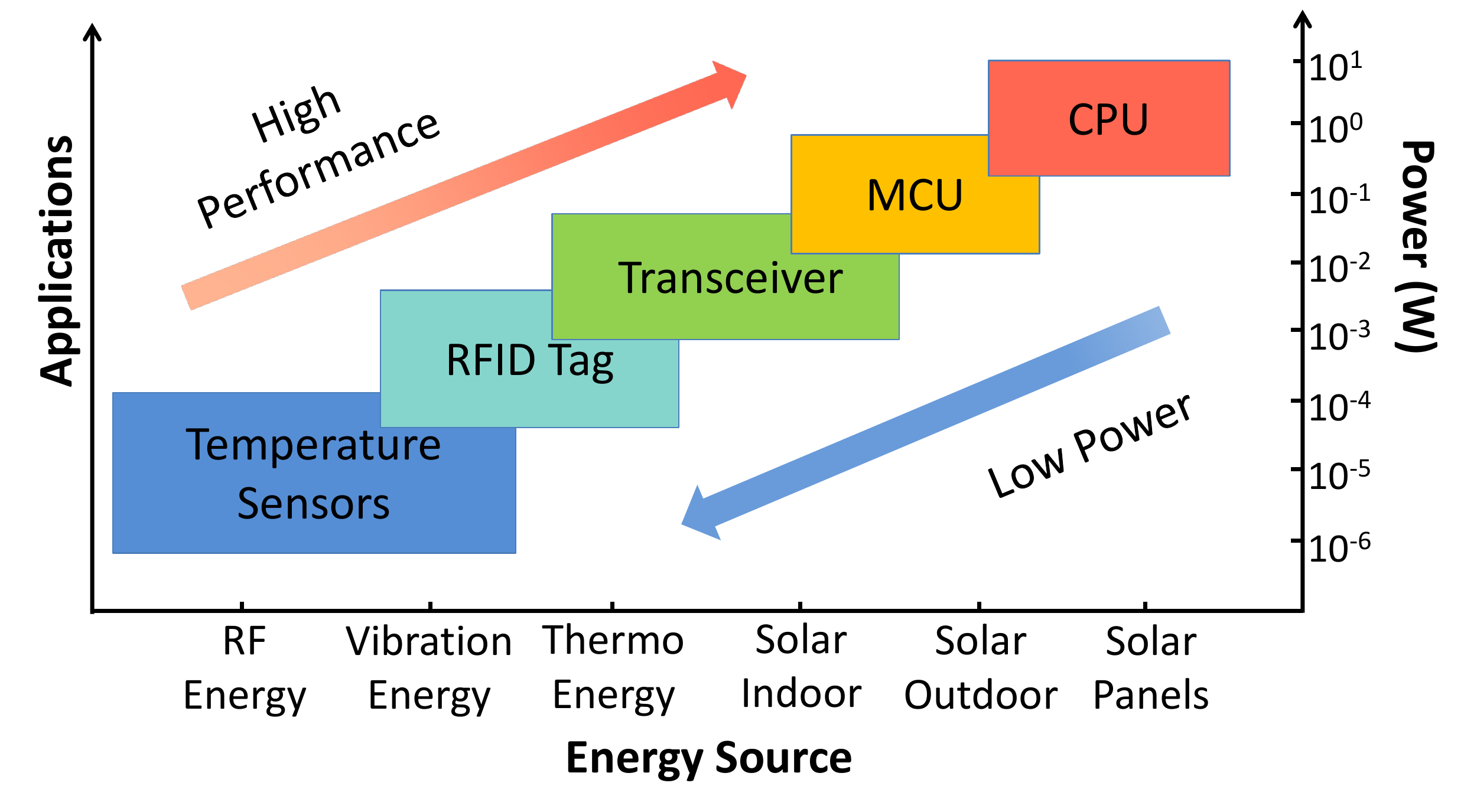}
\caption{The landscape of self-powered electronic devices.}
\label{fig:energy}
\end{figure}

Figure \ref{fig:energy} illustrates the landscape of self-powered sensor nodes and shows the power requirement of various electronic devices and the amount of power that can be harvested by various 
sources like solar energy, thermal, mechanical etc. In particular, for image/video processing and classifications we need high computational power. CPU, GPU and FPGA's are typically used to perform gesture recognition and object classification on video data\cite{gesturefpga, chaofpga}. However, for ``always on'' front-ends where the objective is trigger identification and not continuous gesture recognition, high performance (and hence high power) are not optimal. Instead vision specific MCUs and DSPs are more attractive for self-powered devices, since they exhibit: (1) power dissipation in the order of hundreds of mWs (a 10X reduction compared to CPUs), (2) compact size and low thermal requirements, (3) sufficient computational ability for ``always on'' applications, will be demonstrated here. Our system features an Analog Devices' Black Fin processor. 

To enable ``always on'' and self-powered operation, we take advantage of recent advances in compressed domain (CD) data processing which allows trigger detection with significantly lower power and computational requirements. This is in contrast with existing algorithms which work directly in the pixel domain. Given the objective of our camera front-end, the computation complexity can be largely reduced ($1000 \times$ demonstrated here) from existing algorithms that are targeted for continuous gesture recognition \cite{rautaray2015vision, pavlovic1997visual}. As a command to wake up the system, only a few gesture classes are needed. When the gesture is structured and contains significant motion (for example, writing a big ``Z'' in front of the camera), it can be readily captured by images with high compression ratios. Beyond using low-resolution images, we construct each measurement as a random linear combination of pixels in a manner compatible with compressed domain signal processing. Recent development in compressed sensing and target recognition in the compressed domain \cite{davenport2007smashed, mantzel2012compressive} further improve the accuracy and energy efficiency of the overall process of data acquisition, feature extraction and recognition. We demonstrate that we can take random linear combinations of the pixel vales, and characterize the gesture motion directly from a few compressed measurements. On the other hand, energy harvested from the environment has been used in sensor networks \cite{zhang2013batteryless, liu2015highly} with loads that demand very low power. Here we demonstrate that an algorithm-hardware co-design enables smart camera-front ends with ``always on'' gesture detection. 

The gesture motion is captured by a sequence of difference images between consecutive frames. Each difference image passes two layers of compression to reduce its resolution and to be transferred to the compressed domain. The parameters of the motion are directly extracted from the compressed domain. In section II we describe hardware system architecture \&\ algorithm details in section III. Section IV \&\ V presents hardware implementation \&\ measurement results respectively. Conclusions are drawn in section VI.


\section{Hardware System Architecture}
\begin{figure}[!t]
\centering
\includegraphics[scale=0.31]{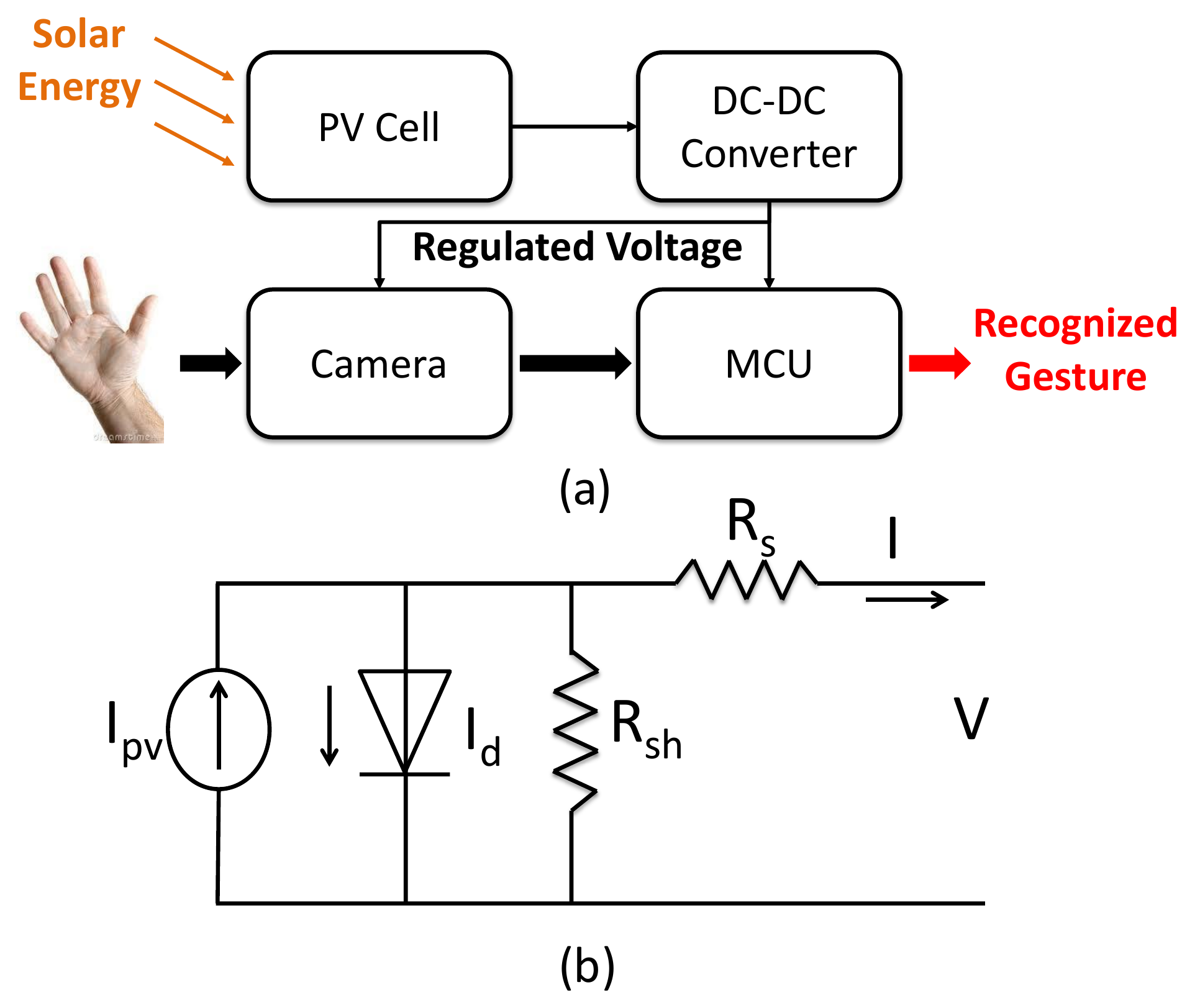}
\caption{(a) Block diagram of the proposed camera front-end, (b) Equivalent circuit representation of a PV cell.}
\label{fig:archi}
\vspace{-0.6cm}
\end{figure}

Before describing the proposed algorithm, we brief the hardware system architecture. The proposed system consists of four main components: a PV cell array, a DC-DC converter with output voltage regulation, an MCU, and an image sensor. The block diagram of our system is shown in Fig. \ref{fig:archi}a. The PV cell converts solar energy to electrical energy. The Norton equivalent output current (Fig.~\ref{fig:archi}b) of PV cell is given by:
\begin{equation}
I=I_{pv}-I_0[exp(\frac{V+IR_s}{aN_sV_T})-1]-\frac{V+IR_s}{R_{sh}} \label{eq:PC_C}
\end{equation}
where I and V are PV cell's output current and voltage respectively; $R_s$ and $R_{sh}$ are the series and shunt resistances; $I_0$, $V_T$, $a$, $N_s$ are dark saturation current,  thermal voltage, diode ideality factor, and number of cell connected in series respectively; $I_{pv}$ is the generated current whose magnitude depends on irradiation and temperature.

\begin{figure}[!b]
\centering
\includegraphics[scale=0.38]{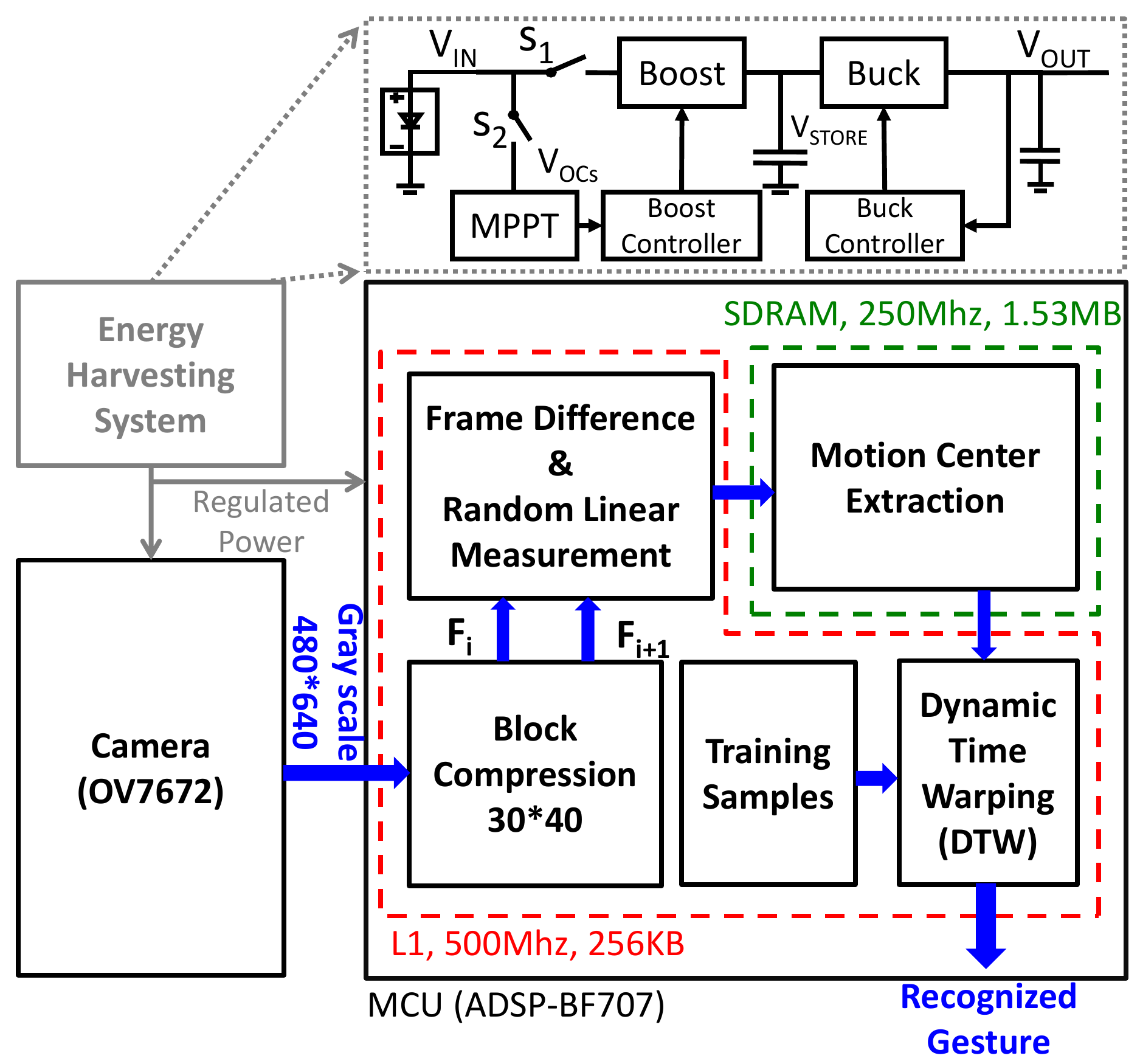}
\caption{Hardware system architecture demonstrating key components of power management, the MCU and the mapping of the algorithm on the MCU.}
\label{fig:BQ25570}
\end{figure}

As the MCU and image sensors both demand regulated voltage to operate, the DC voltage generated by PV cells must be regulated by a DC-DC converter. For the current design, we select TI's BQ25570EVM, a two-stage DC-DC converter with Maximum Power Point Tracking (MPPT) for solar energy harvesting and for providing a regulated output supply. The block diagram of the energy harvesting system and gesture recognition flow is shown in Fig. \ref{fig:BQ25570}.

The input image is captured by Omnivision's OV7672 sensor with a native resolution of $480 \times 640$. We extract only the gray-scale component of the image, which reduces the computation power without any impact on performance. The output of the pixel array is passed on to an on-board ADSP BF707 MCU using I2C interface. Once the image is received by the BF707 processor, we perform the following operations: block averaging, frame difference, random linear measurements, motion centers extraction in compressed domain followed by gesture recognition. For block compression we extract every one out of 16 pixel values in each row and column. Therefore the block compression factor is 256 (16 for every row and column). The block average, frame difference and dynamic time warping related matrices are stored in L1 cache (requires less than 128KB). Motion center extraction co-efficients are stored in external SDRAM (requires more than 1.2MB). L1 access is performed using core clock at 500MHz and SDRAM access happens at system clock with 250MHz speed. The hardware is further optimized by (1) using short integer maths and (2) optimizing memory usage that reduces total power consumption without loss of performance. 

\section{Gesture recognition algorithm}
\begin{figure}[t!]
\centering
\includegraphics[scale=.48]{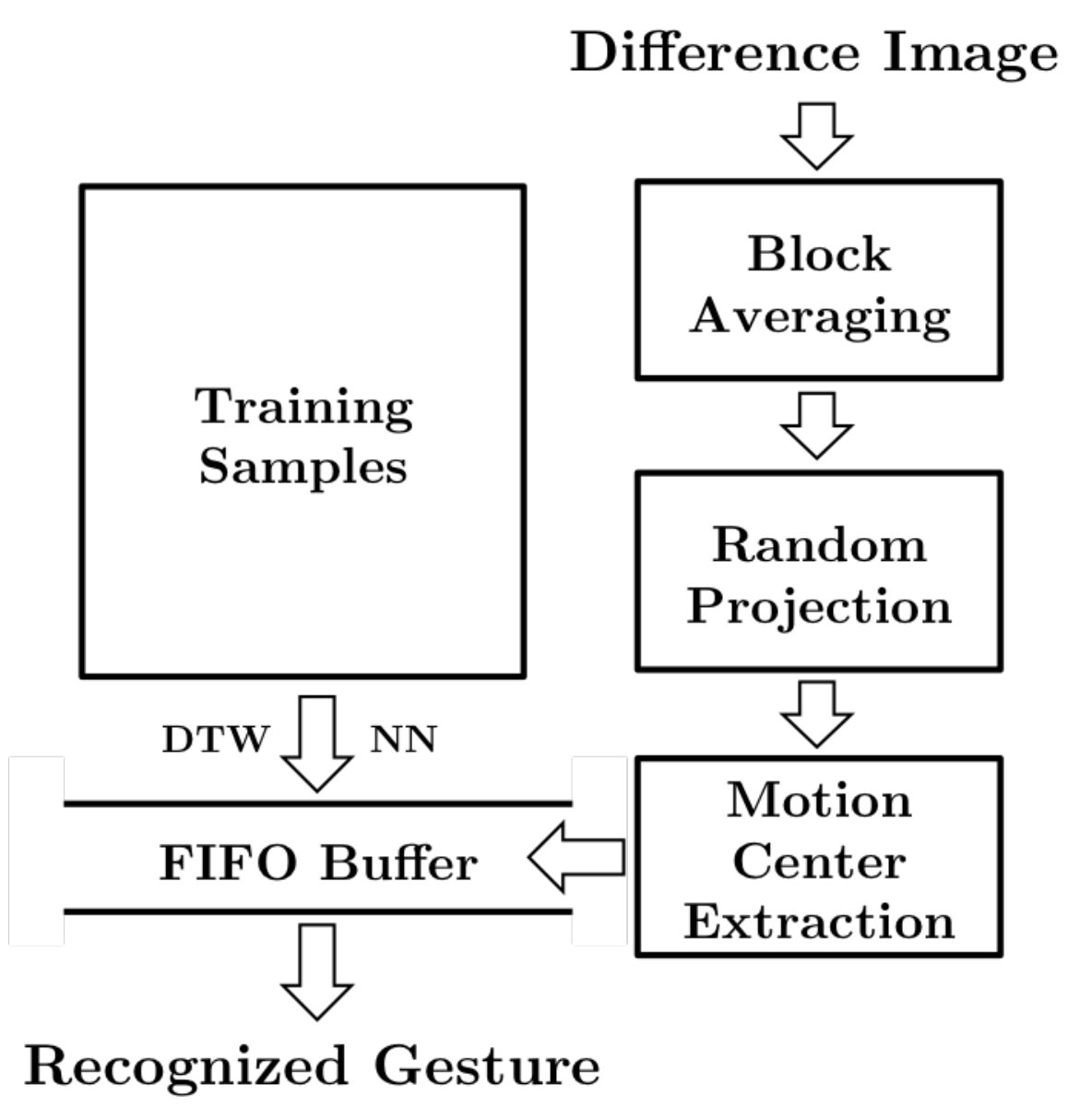} 
\caption{Block diagram of the proposed algorithm.} \label{Block_Diagram}
\vspace{-0.6cm}
\end{figure}

Our real-time gesture recognition algorithm is based on motion parameters extracted directly from the compressed domain. The starting point of our algorithm is the difference image. When a user's hand is the only significant moving object present in front of the camera, the hand region is well presented by the difference image which then passes through two layers of compression. In the first layer, the resolution is reduced by dividing the whole image into several blocks and taking the average of each block. In the second layer, we transfer this low resolution image to the compressed domain by taking random linear combinations of its pixels. We estimate the center of the motion directly in the compressed domain without recovery the image sequences. These motion centers are passed to a nearest neighbor (NN) classifier coupled with DTW distance measurement for gesture recognition. The block diagram of our system is shown in Fig. \ref{Block_Diagram}.

\begin{figure*}[h!]
\centering
\begin{subfigure}[b]{0.28\linewidth}
\centering
\includegraphics[width=\linewidth]{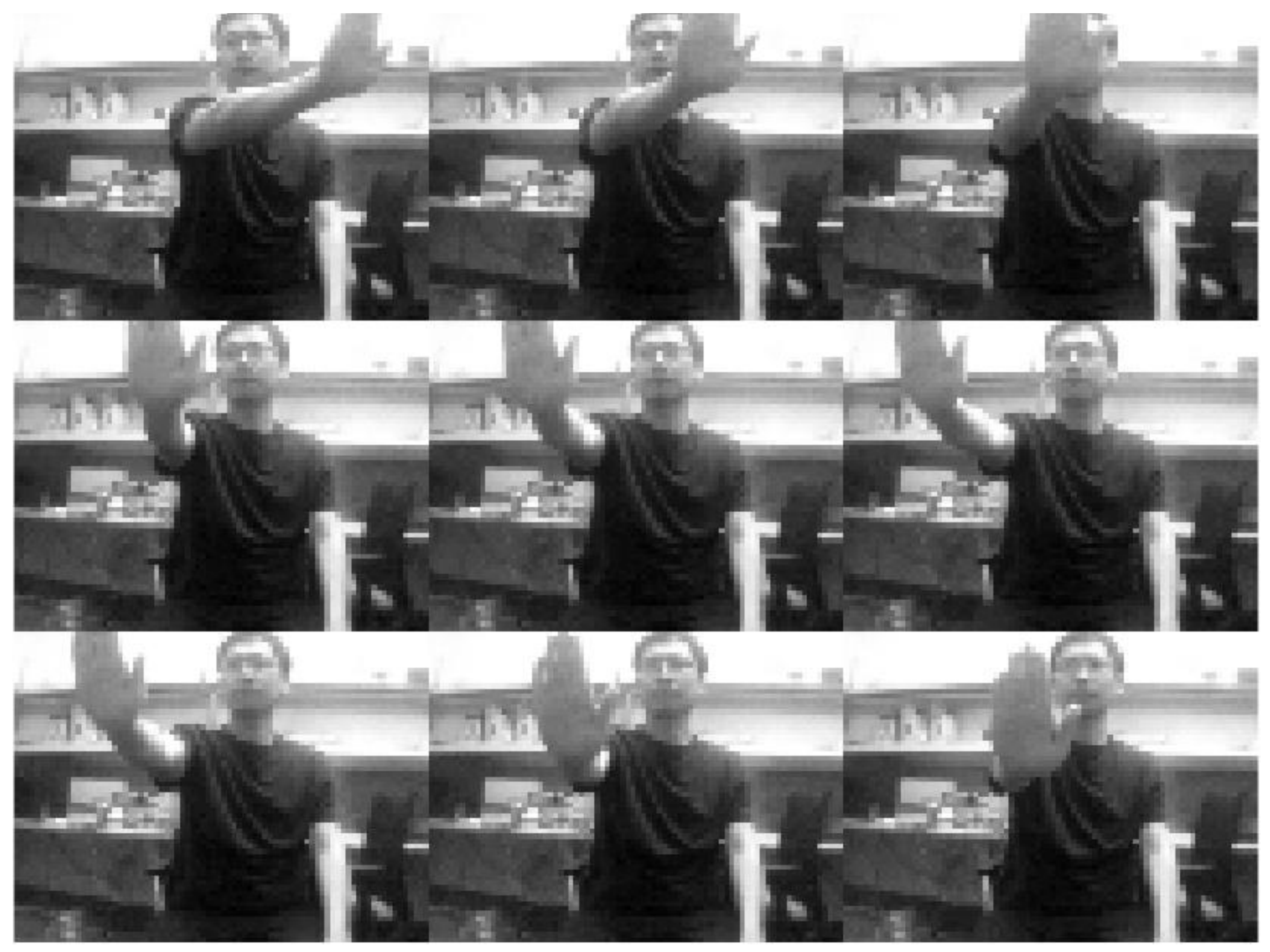}
\caption{}
\label{fig:gesture}
\end{subfigure}
\begin{subfigure}[b]{0.28\linewidth}
\includegraphics[width=\linewidth]{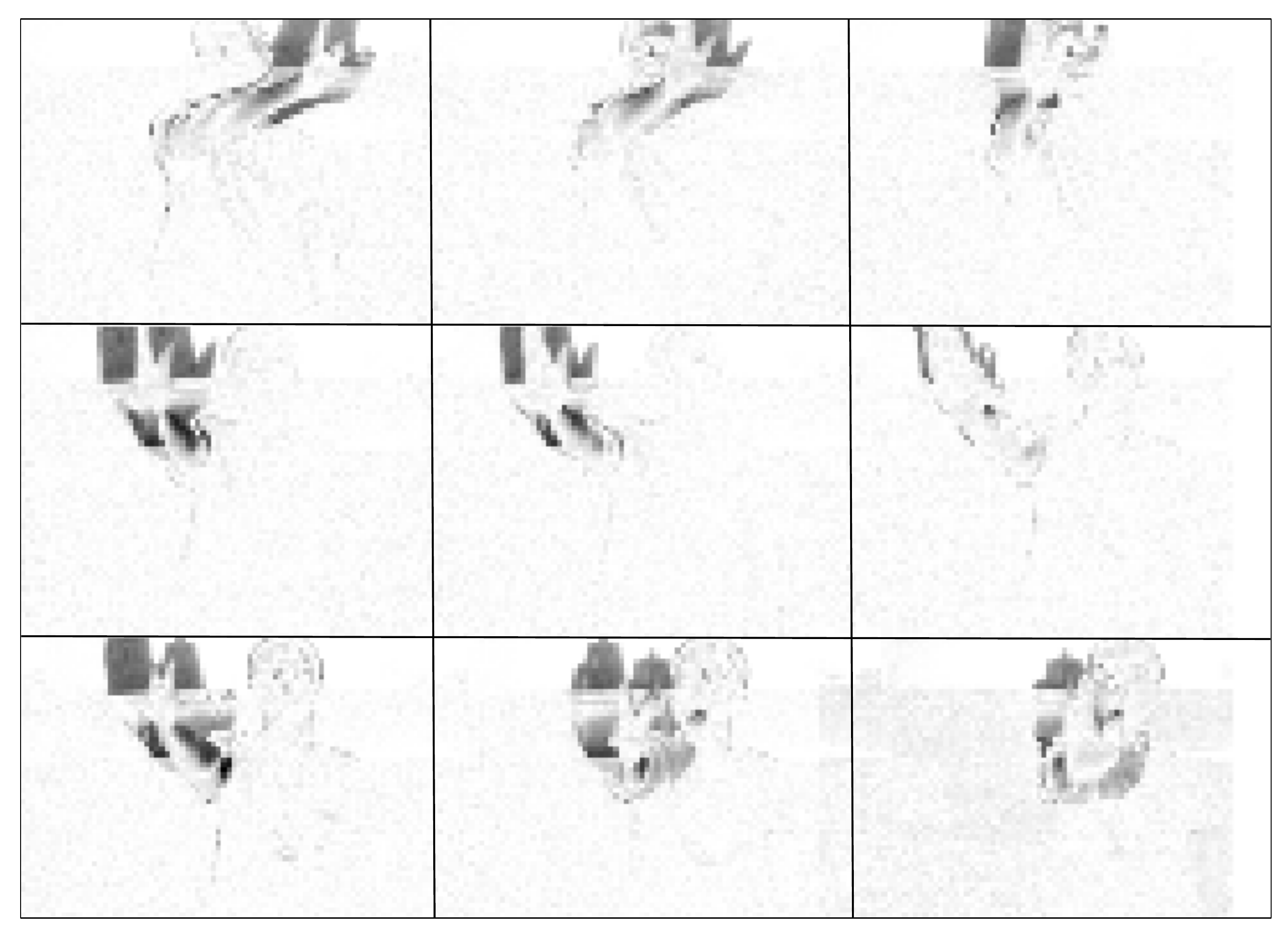}
\caption{}
\label{fig:frame_diff}
\end{subfigure}
\begin{subfigure}[b]{0.28\linewidth}
\includegraphics[width=\linewidth]{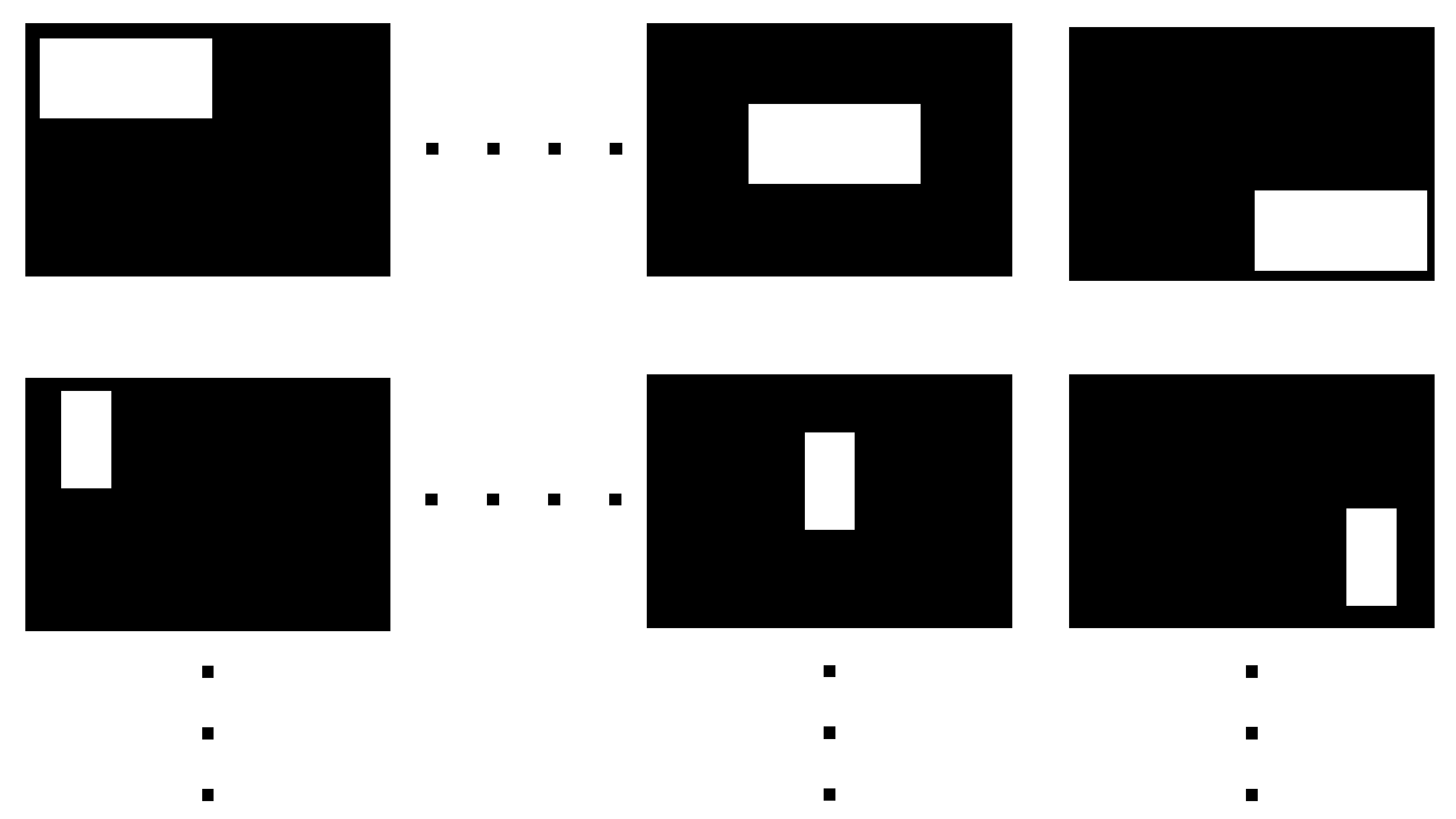}
\caption{}
\label{fig:template}
\end{subfigure}
\caption{Measurement results illustrating (a) a sequence of images captured by the camera, (b) the computed frame differences, and (c) Matching templates in the uncompressed domain. Rectangle sizes differ among rows and centers of the rectangles differ among columns}
\label{fig:frames}
\vspace{-0.6cm}
\end{figure*}

\subsection{Two layers of compression}
We denote $F_i$ as the $i$th full resolution image output from the camera with size($W$ by $H$). The difference image $D_i$ (Figure \ref{fig:frame_diff}) of two consecutive frames (Fig. \ref{fig:gesture}) is given by:
\begin{equation}
D_i=|F_{i+1}-F_i|
\end{equation}
In the first compression layer, the difference image is divided evenly into blocks of size $B \times B$. The average of the pixel values in each block is taken, resulting in a block compressed difference image of size $\frac{W}{B} \times \frac{H}{B}$ 
We vectorize this low-resolution difference image and denote it as $y_i \in \mathbb{R}^N$. In the second layer of compression, we construct a random matrix $\Phi$ of size $M \times N$. Each entry of $\Phi$ is uniformly chosen from $\{+1, -1\}$. The projection of the vectorized low-resolution difference image in the compressed domain is calculated as:
\begin{equation}
\hat{y}_i = \Phi y_i 
\end{equation}
Each entry in $\hat{y}_i \in \mathbb{R}^M$ is a random linear combination of all the entries in $y_i$. We can write both compression layers into one linear equation:
\begin{equation}
\hat{y}_i = \Theta Y_i = \Phi\Psi Y_i \label{eq:cf}
\end{equation}
Where $Y_i$ is the vectorized original difference image $D_i$. $\Psi$ is the block averaging matrix of size $N$ by $W\times H$. Its product with $\Phi$ forms a structured random matrix $\Theta$.

\subsection{Motion center extraction in the compressed domain}
In the uncompressed low-resolution domain, the hand region in the difference image can be captured by a template shown in Figure \ref{fig:template}. The template (of size $\frac{W}{B}$ by $\frac{H}{B}$) has uniform non-zero values within the small rectangular region and is zero elsewhere. To locate the hand region, we construct a set of vectorized templates $X(\alpha , r)$, where $\alpha$ represents the coordinates of the center of the small rectangle, and $r$ represents different rectangle sizes. The variation in sizes is to adapt to the change of the hand size seen by the camera when users are at different locations. The center of the hand motion is extracted by solving
\begin{equation}
(\alpha^*, r^*) = \argmin{\alpha, r} ||y_i - X(\alpha , r)||_2 \label{eq:uc}
\end{equation}
The collection of templates forms a manifold in $\mathbb{R}^N$ with intrinsic parameters $\alpha$ and $r$. Using the result from \cite{baraniuk2009random} and \cite{davenport2007smashed}, we can directly extract the motion centers in the compressed domain. That is, for
\begin{equation}
(\hat{\alpha}^*, \hat{r}^*) = \argmin{\alpha, r} ||\hat{y}_i - \Phi X(\alpha , r)||_2 \label{eq:c1}
\end{equation}
$(\hat{\alpha}^*, \hat{r}^*) \approx (\alpha^*, r^*)$ with high probability for some $M << N$. By choosing the norm as $L_2$ and normalizing all the templates to have the same energy, we can further write motion center estimation as solving:
\begin{equation}
(\hat{\alpha}^*, \hat{r}^*) = \argmax{\alpha, r} \hat{y}_i^T \Phi X(\alpha , r) \label{eq:c2}
\end{equation}

The above process represents a smashed filter operation which is akin to matched filters but in the compressed domain. Since we are not interested in reconstructing the image from compressed measurements, a compressed domain smashed filter reduces the computation by ($M/N$).

\subsection{Gesture Recognition}
The extracted motion centers are stored in a FIFO buffer with length $L$. We then measure the distances between the latest data in the buffer and the training samples. As gestures can be performed with different speed, the sequences of the gestures' motion centers are of different lengths. We implement DTW algorithm, which automatically extracts the best matching segments between two sequences, adjusting them to the same length, and finally calculating the distance\cite{muller2007dynamic}. Once the smallest distance passes a threshold, we assign the gesture to the class that has its nearest sample. Algorithm performance on the proposed hardware will be described in Section V.

\section{Experimental setup}

\subsection{Power Management Design}

The overall platform is designed from COTS components and here we explain the optimal design choice. We chose omnivision OV7672 image sensor which has frame size of $480 \times 640$ pixels. The image sensor is connected to an ADSP BF707 processor using I2C interface. Measurements reveal a maximum current consumption of $170mA$ at fixed $3.3V$ power supply. 

\begin{figure}[b!]
\centering
\begin{subfigure}[b]{0.42\linewidth}
\centering
\includegraphics[width=\linewidth]{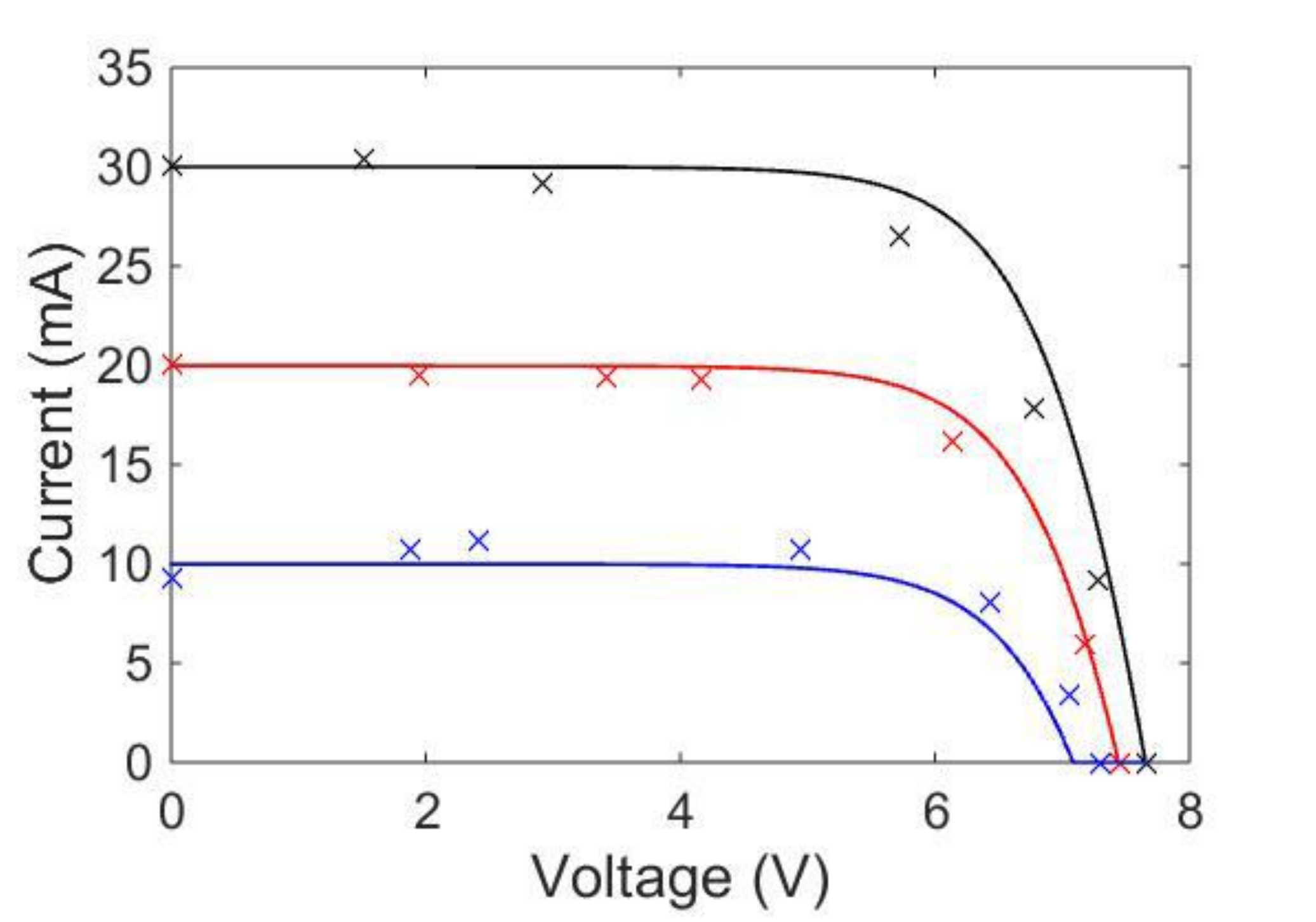}
\caption{I-V characteristics}
\label{fig:PV_IV}
\end{subfigure}
\begin{subfigure}[b]{0.42\linewidth}
\includegraphics[width=\linewidth]{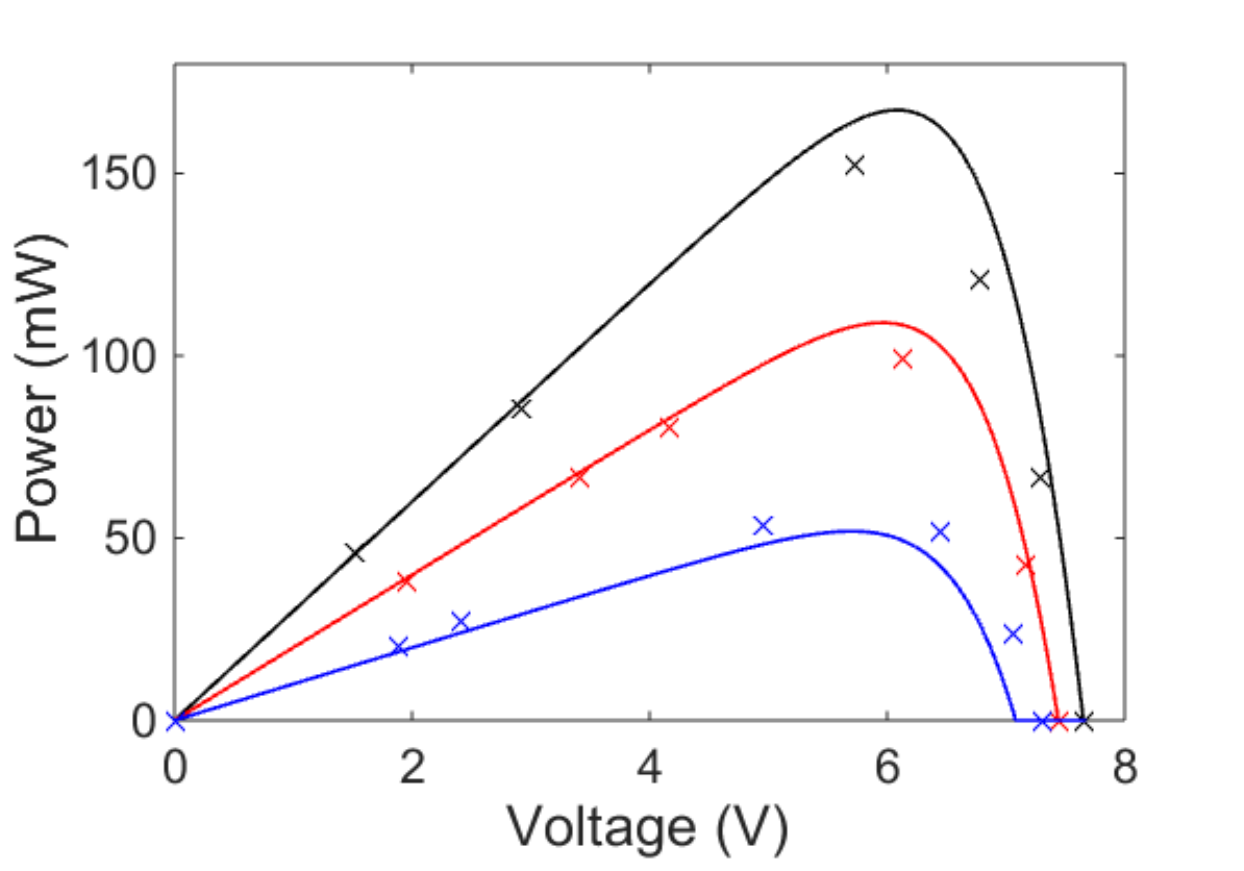}
\caption{P-V characteristics}
\label{fig:PV_PV}
\end{subfigure}
\caption{ (a) I-V characteristics and (b) Power-Voltage characteristics of PV cells at three different irradiance levels (300W/$m^2$, 600W/$m^2$ and 1000W/$m^2$). Discrete points are experimental results and continuous curves are from simulations.}
\end{figure}

The solar cell (AM5907) produces an output voltage of 5V at the point of maximum power transfer. The I-V and P-V characteristics of each cell is shown in Fig. \ref{fig:PV_IV} and Fig. \ref{fig:PV_PV} vis-a-vis simulation results. We see a close match between experimental results and empirically fitted Eqn~\ref{eq:PC_C}. We note that for an irradiance of $600W/m^2$, the maximum power $\approx 100mW$. In the current setup, We use 6 PV cells in parallel to generate the required power that the load demands. Also, from Figure \ref{fig:PV_PV}, we observe that operating voltage at maximum power point is approximately $80\%$ of the open circuit voltage ($V_{OC}$). Hence Maximum Power Point Tracking (MPPT) is achieved by regulating the output at $80\%$ of $V_{OC}$. 

\begin{figure}[t!]
\centering
\begin{subfigure}[b]{0.44\linewidth}
\centering
\includegraphics[width=\linewidth]{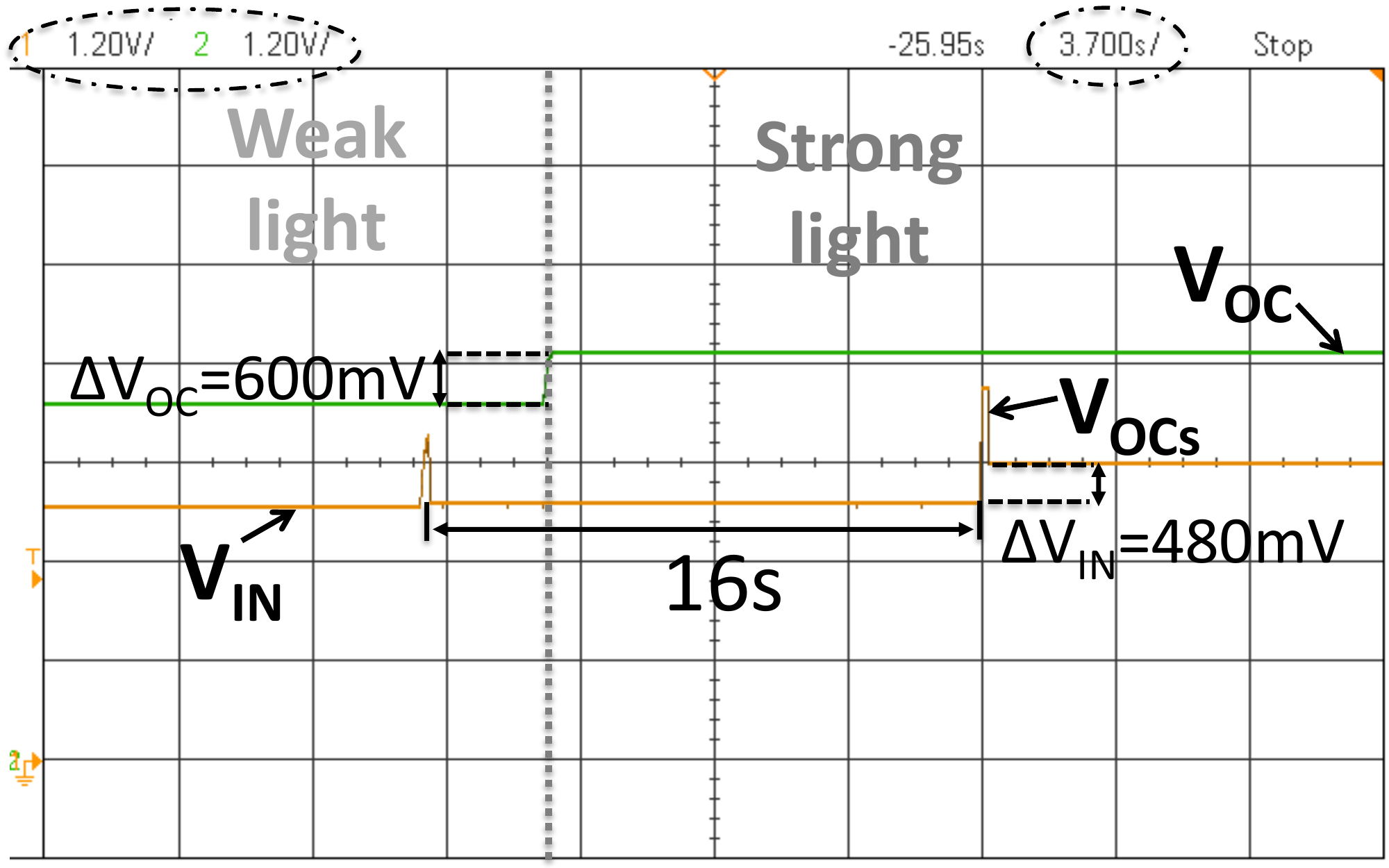}
\caption{}
\label{fig:MPPT}
\end{subfigure}
\begin{subfigure}[b]{0.44\linewidth}
\includegraphics[width=\linewidth]{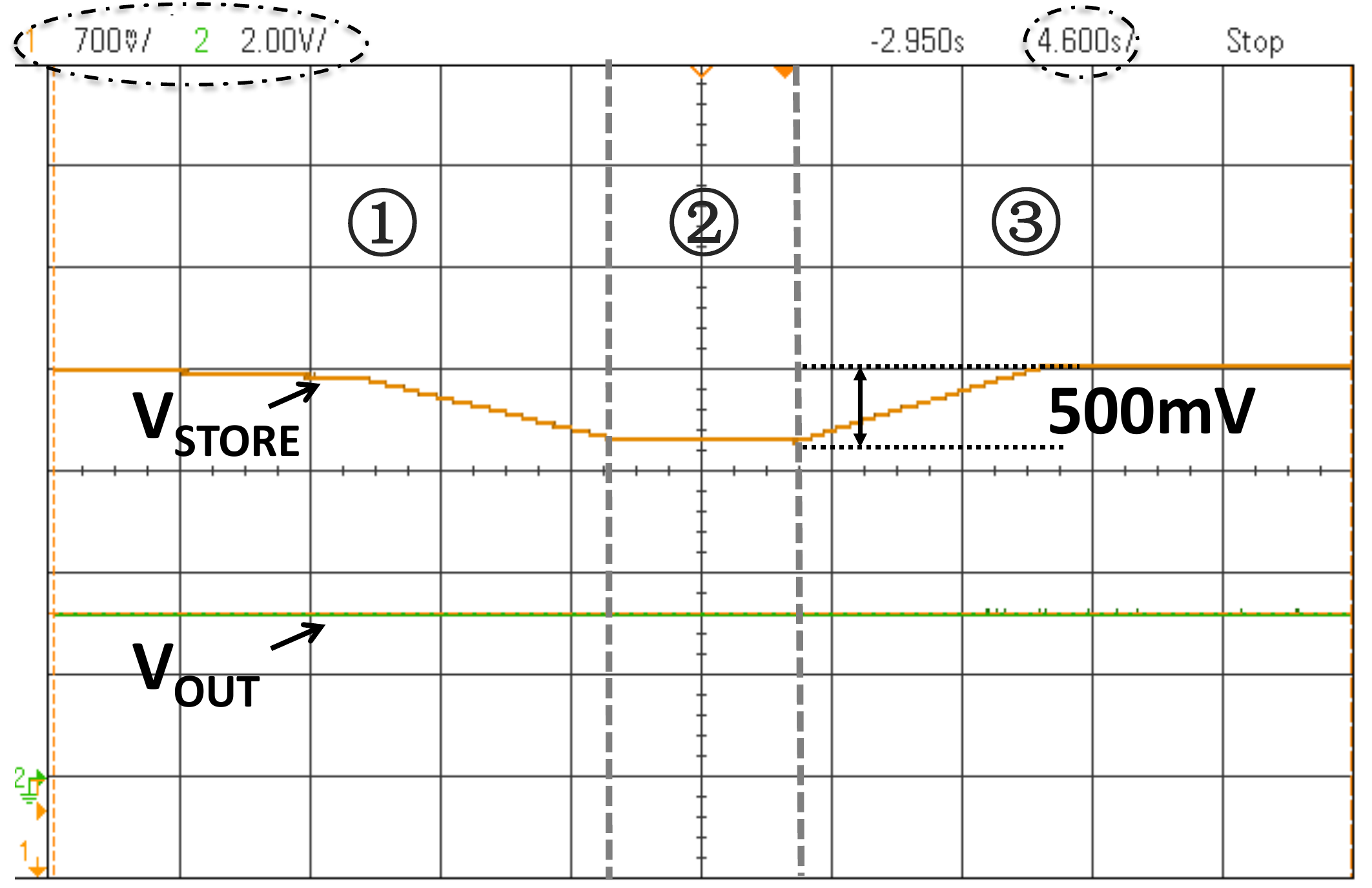}
\caption{}
\label{fig:VSTORE}
\end{subfigure}
\caption{Oscilloscope captures illustrating (a) MPPT where $V_{IN}$ tracks the open circuit voltage at $80\%$ of $V_{OC}$ as irradiance changes and (b) regulation of $V_{OUT}$ under dynamically varying super-capacitor
voltage ($V_{STORE}$). In (b) three regions are shown: (1) instantaneous load power consumption is higher than input power which reduces $V_{STORE}$; (2) load power and the harvested power are balanced and (3) load consumption is less than harvested power.}
\label{fig:fig1}
\vspace{-0.6cm}
\end{figure}

\begin{figure}[!b]
\centering
\includegraphics[scale=0.43]{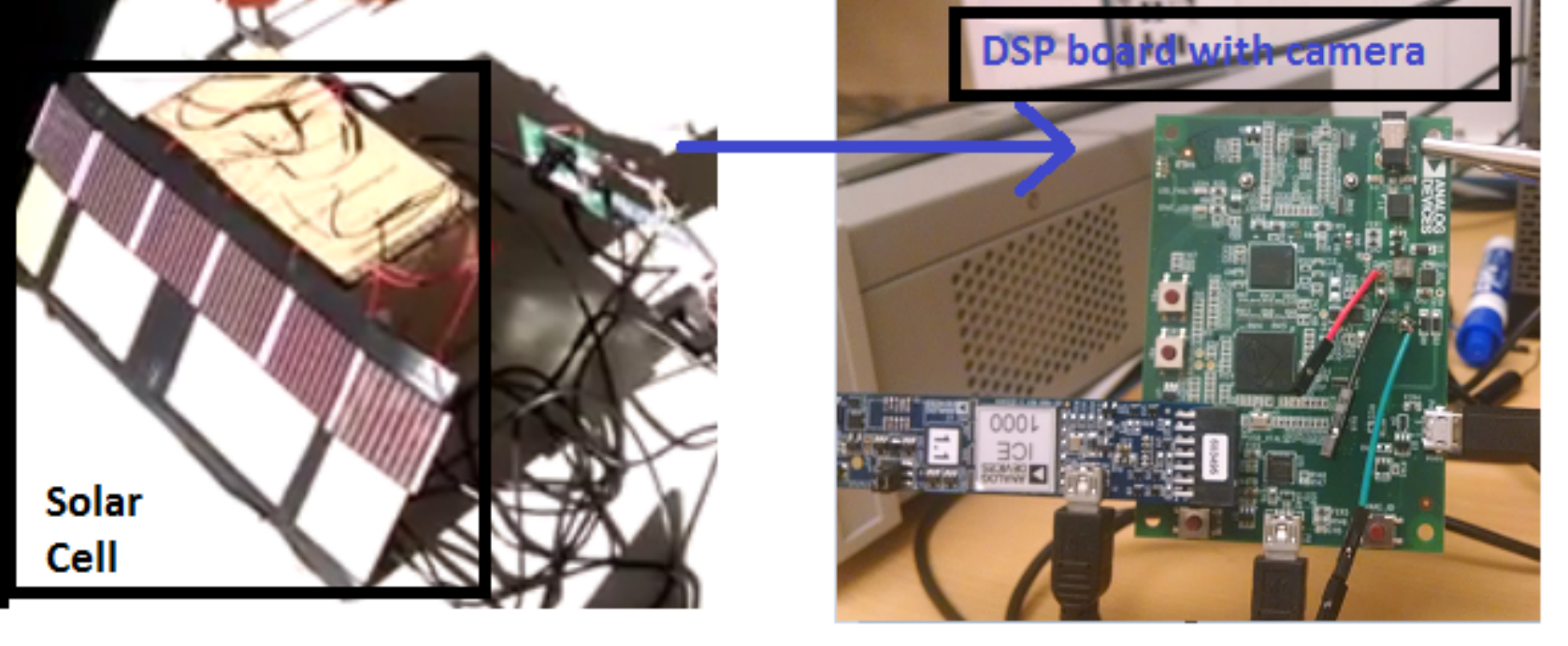}
\caption{The overall system demonstrating the solar cells and the MCU with the camera.}
\label{board_full}
\end{figure}

As shown in Fig. \ref{fig:BQ25570}, the MPPT block samples open circuit voltage $V_{OC}$ every 16 seconds with $S_2$ on and $S_1$ off. This sample voltage $V_{OC\_sample}$ is sent to the boost controller to modulate the phase and frequency of the boost converter so that the PV cell operates at maximum power point, $80\%$ of $V_{OC\_sample}$. The sampling process is shown in oscilloscope captures in Fig. \ref{fig:MPPT}. It is observed that open circuit voltage is sampled and the PV cell's operating voltage changes accordingly. The energy is stored in a super-capacitor between the two converter stages. Availability of super-capacitor benefits camera-based applications whose power requirement fluctuates significantly. The output voltage is sensed and sent back to buck controller to regulate the output voltage. The output voltage is hardware programmable through programmable external resistors on the board. Fig. \ref{fig:VSTORE} shows how $V_{STORE}$ varies with varying irradiance and load current conditions. Measured oscilloscope capture also reveals that $V_{OUT}$ is well regulated under such dynamic conditions. The complete experimental setup along with the PV cells and the MCU is shown in Fig. \ref{board_full}.

\begin{figure*}[!t]
\centering
\includegraphics[scale=.61]{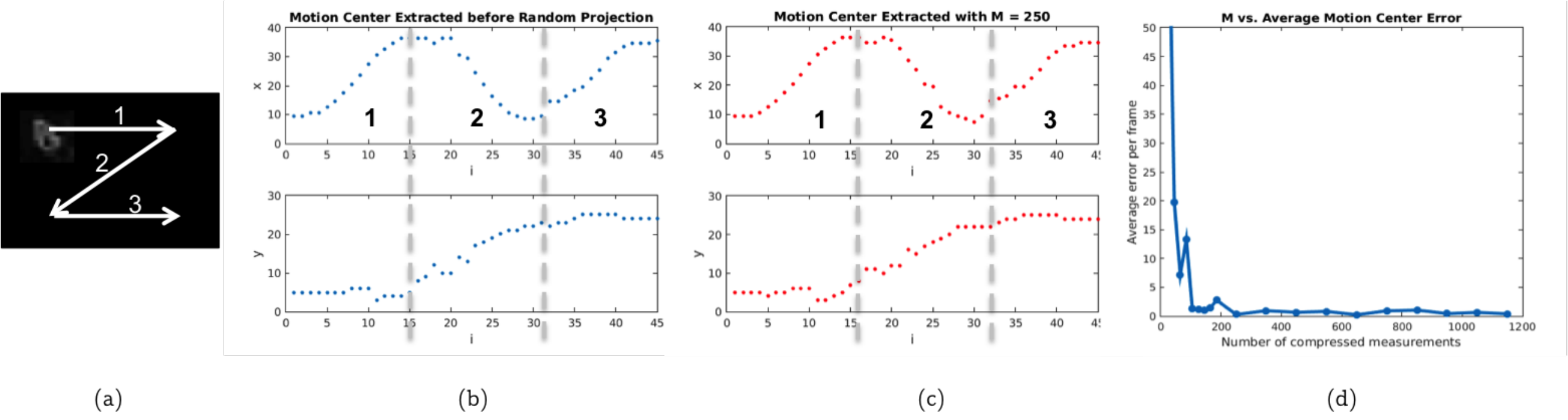}
\caption{Simulation results (from MATLAB) for different numbers of compressed measurements $M$. (a) Hand motion of gesture "Z" divided into three segments, (b) Motion center extracted before random projection by solving equation (\ref{eq:uc}), (c) Motion center extracted from 250 compressed measurements by solving equation (\ref{eq:c2}), (d) Average error of motion center extracted in the compressed domain compared with (b).} \label{M_vs_MC}
\vspace{-0.6cm}
\end{figure*}

\subsection{Mapping Proposed Gesture Recognition Algorithm on Low-Power MCU}
The image sensor output at full-resolution $(480\times 640)$ is captured by the MCU. The MCU performs compression on each difference image. In the block compression layer, we choose blocks of size $16 \times 16$; and hence the vectorized low-resolution difference image  $y_i \in \mathbb{R}^{1200}$. The compression rate of this layer is thus $256$. 
In the random projection layer, the number of compressed measurements $M$ is a design variable. To gain better insights on the choice of $M$, we explore its relationship with the accuracy of motion center extraction. For a typical gesture "Z" the extraction algorithm is shown in Figure \ref{M_vs_MC}a. The motion centers are extracted from the block-averaged difference images by solving equation (\ref{eq:uc}). As we can see in Figure \ref{M_vs_MC}b, the three segments of the gesture are clearly distinguished on the path of the motion centers. With $M = 250$, the motion centers are extracted in the compressed domain by solving equation (\ref{eq:c2}), and are plotted in Figure \ref{M_vs_MC}c. The similarity between this plot and \ref{M_vs_MC}b demonstrates the effectiveness of the theory. For each value of $M$ we calculate the average motion center error per frame in the compressed domain. The ``L'' shape of the curve indicates that $M=250$ is the threshold for nearly error-free motion parameter estimation, granting us another factor of 5 compression rate. This ``threshold'' behavior is consistent with the classic results from compressed sensing presented in~\cite{baraniuk2009random,davenport2007smashed,mantzel2012compressive}. The accurate motion center extraction in the compressed domain provides the foundation of preserving high recognition accuracy.
\vspace{-0.1cm}

To reduce memory usage and reduce power consumption, we fix the size of the smashed filter templates (Figure \ref{fig:template}) to $10 \times 10$. In other words, we construct $X(\alpha , r)$ with $r$ fixed to $10 \times 10$ and $\alpha$ being every possible location in the $40 \times 30$ block-averaged difference image. Using the same $\Phi$, we transfer all the templates into the compressed domain by calculating $\Phi X(\alpha , r)$. 
\vspace{-0.1cm}

As proof of concept, we tested the system with a variety of key gestures and in the rest of the paper, we will discuss an implementation that recognizes $3$ gesture classes: "X", "+", and "Z". For the usage model where the key gestures are used for ``wake up'', a small number of gesture classes suffices. In each class, we provide $40$ training examples. In each training example, the gesture is performed at different locations with respect to the camera, and the motion centers were extracted from the uncompressed domain by solving equation (\ref{eq:uc}). For low power operation and to enable a completely, self-powered system, the image sensor is operated at a maximum of $10$ frames/second and the buffer length is set to $50$.

\section{Measurement Results}

\begin{figure}[!h]
\centering
\includegraphics[scale=.41]{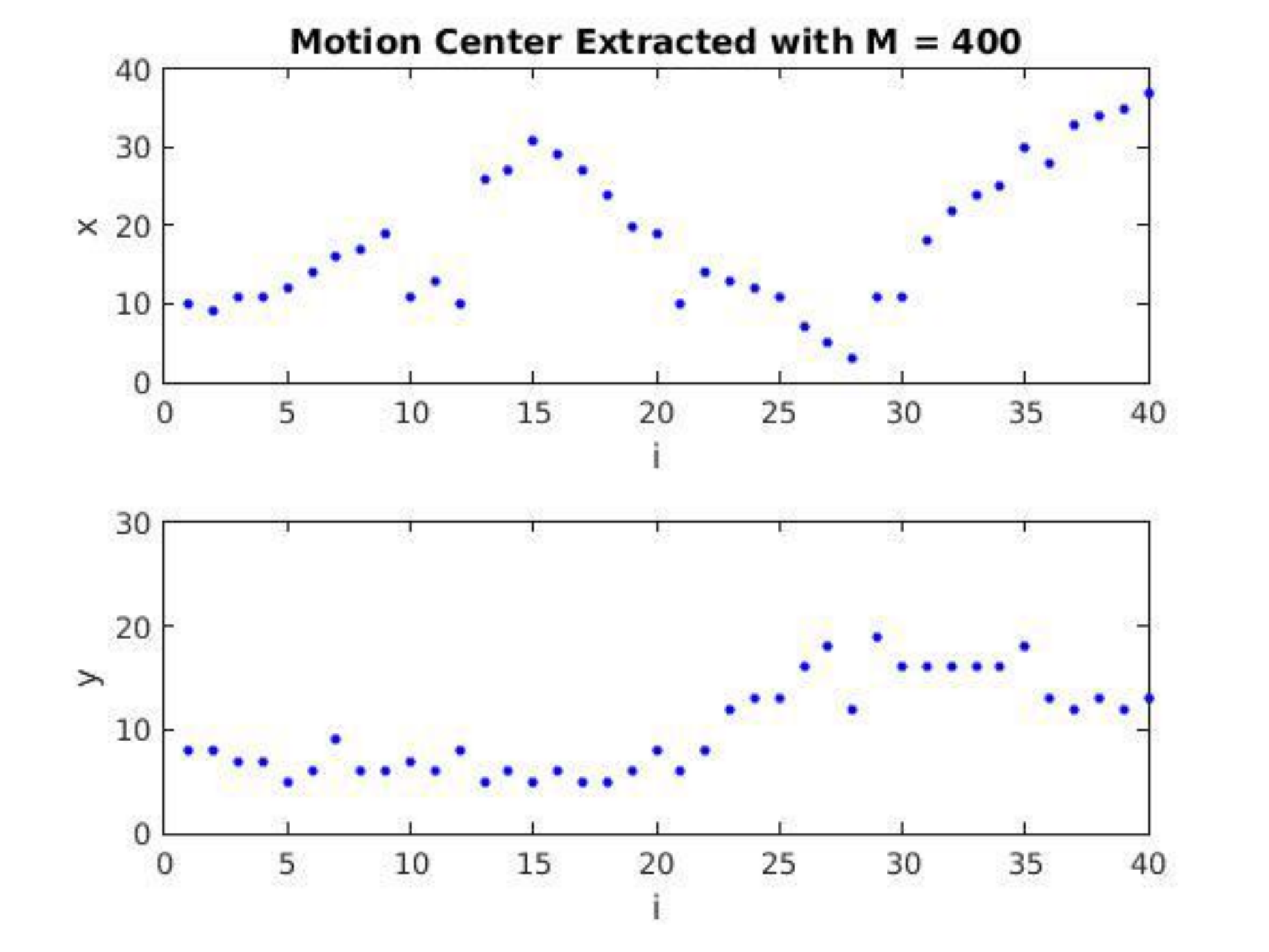}
\caption{Measured motion center extraction of hand gesture ``Z'' for $M=400$(Close match with simulations in Fig.~\ref{M_vs_MC}.} 
\label{Zt}
\vspace{-0.7cm}
\end{figure}

\subsection{M vs. Recognition Rate and Power Consumption}

\begin{figure*}[t!]
\centering
\begin{subfigure}[b]{0.31\linewidth}
\centering
\includegraphics[width=\linewidth]{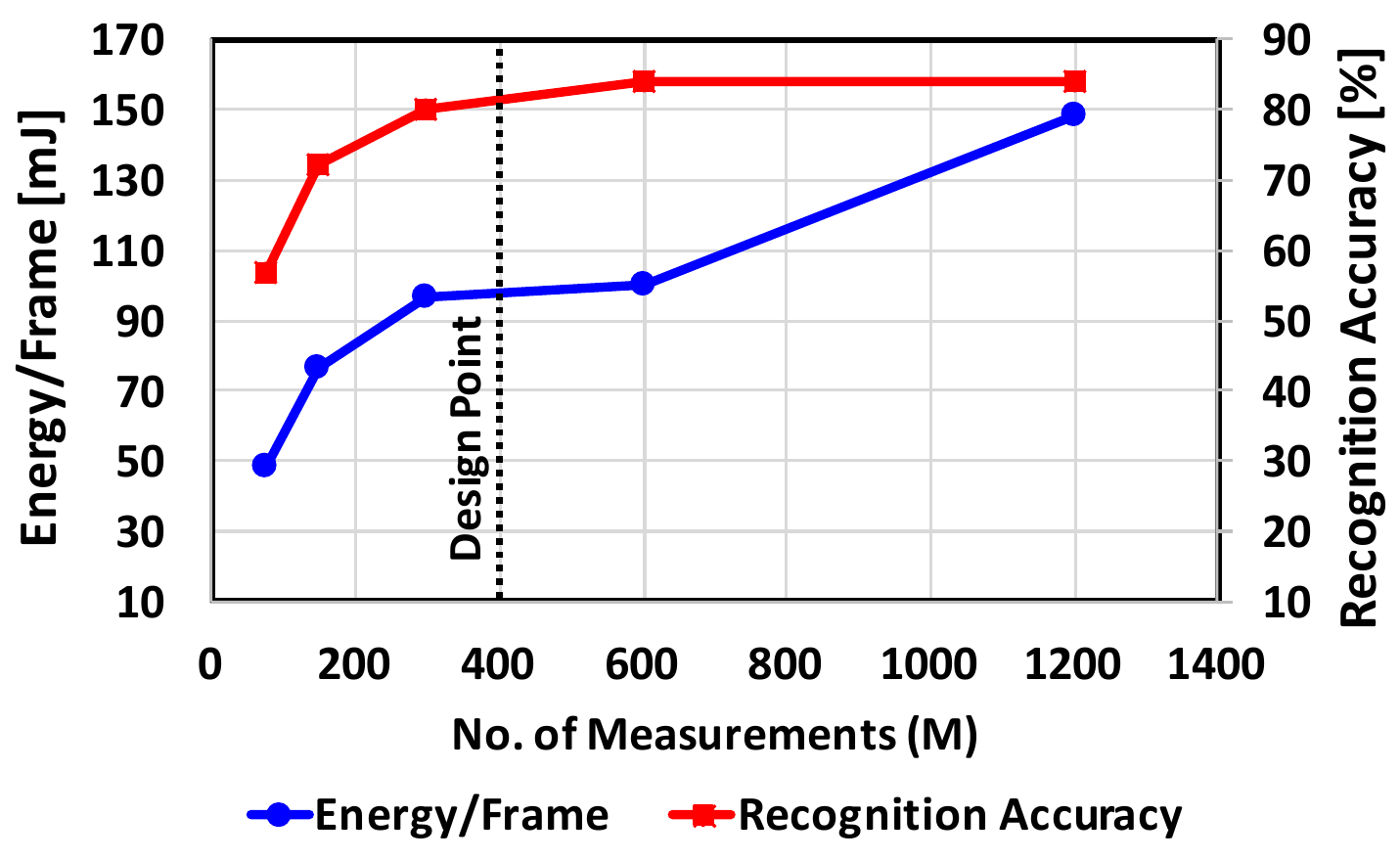}
\caption{}
\label{fig:power_nmeas}
\end{subfigure}
\begin{subfigure}[b]{0.31\linewidth}
\includegraphics[width=\linewidth]{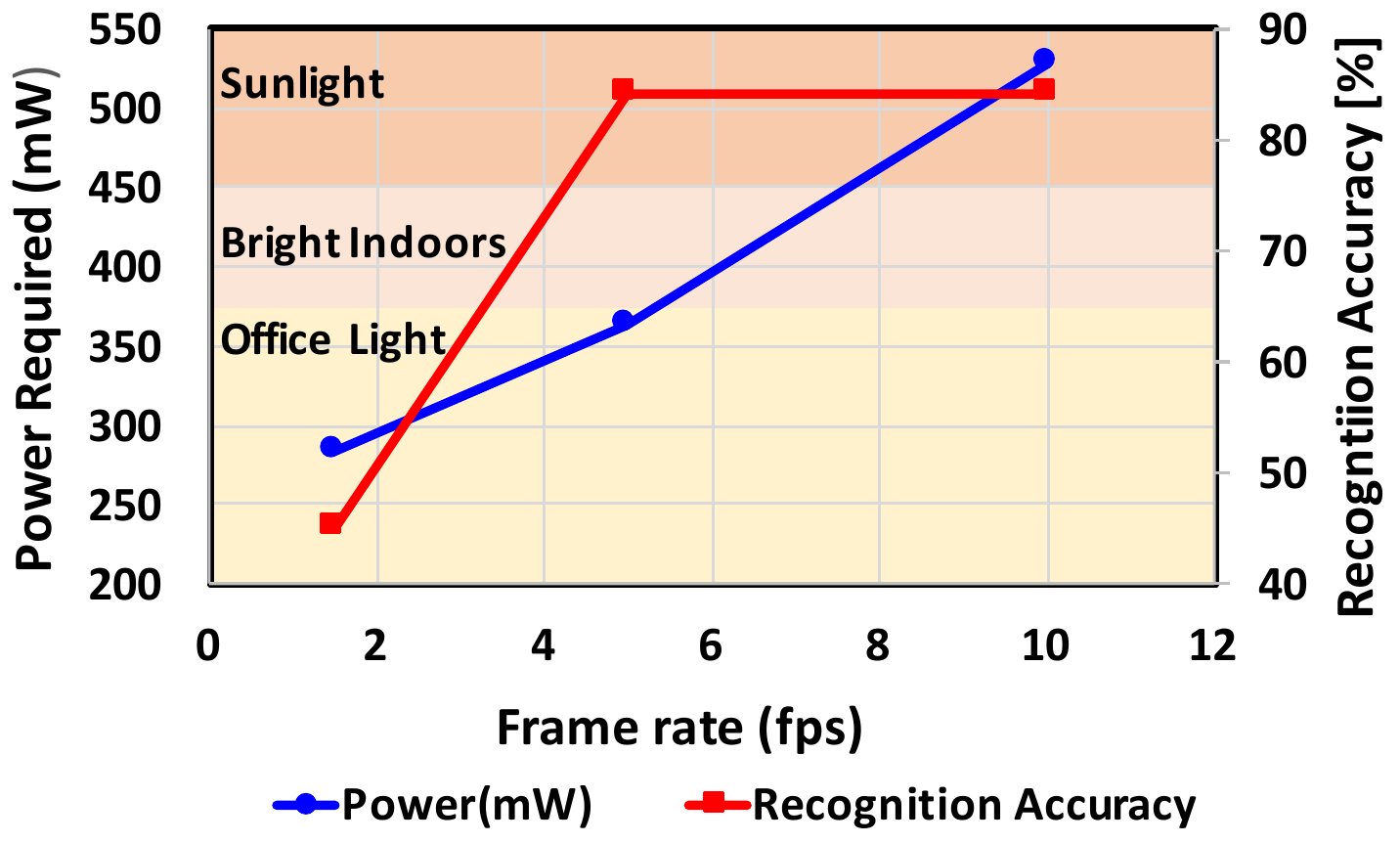}
\caption{}
\label{fig:power_frame}
\end{subfigure}
\begin{subfigure}[b]{0.31\linewidth}
\includegraphics[width=\linewidth]{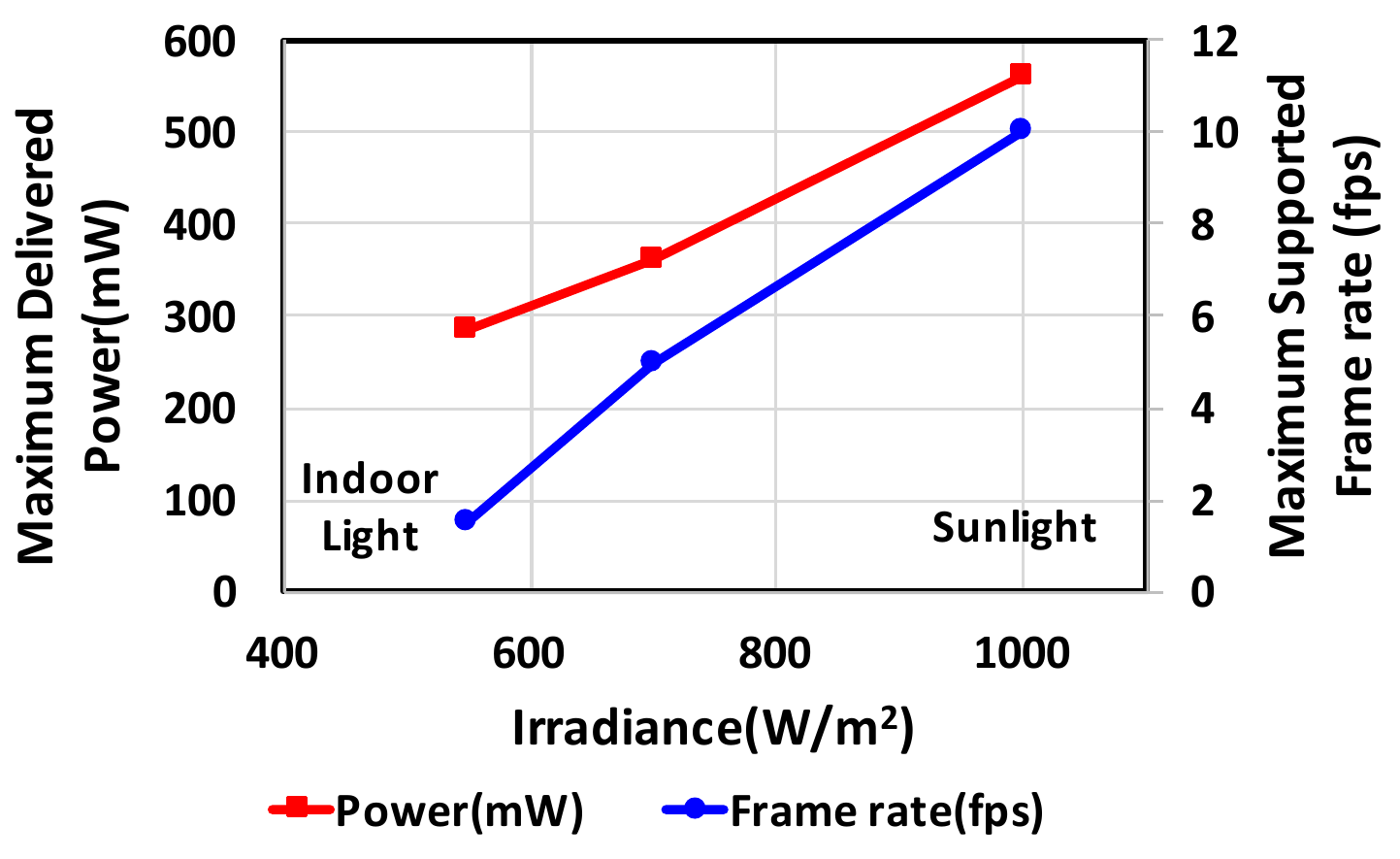}
\caption{}
\label{fig:Irradiance}
\end{subfigure}
\caption{Design space exploration through measurements: (a) Number of compressed measurements (M) vs. Energy/frame and Recognition accuracy, (b) Frame rate vs. Power and Recognition accuracy, and (c) Frame rate and Power vs Irradiance.}
\label{fig:power_performance}
\vspace{-0.7cm}
\end{figure*}

\begin{figure}[!b]
\centering
\includegraphics[scale=0.4]{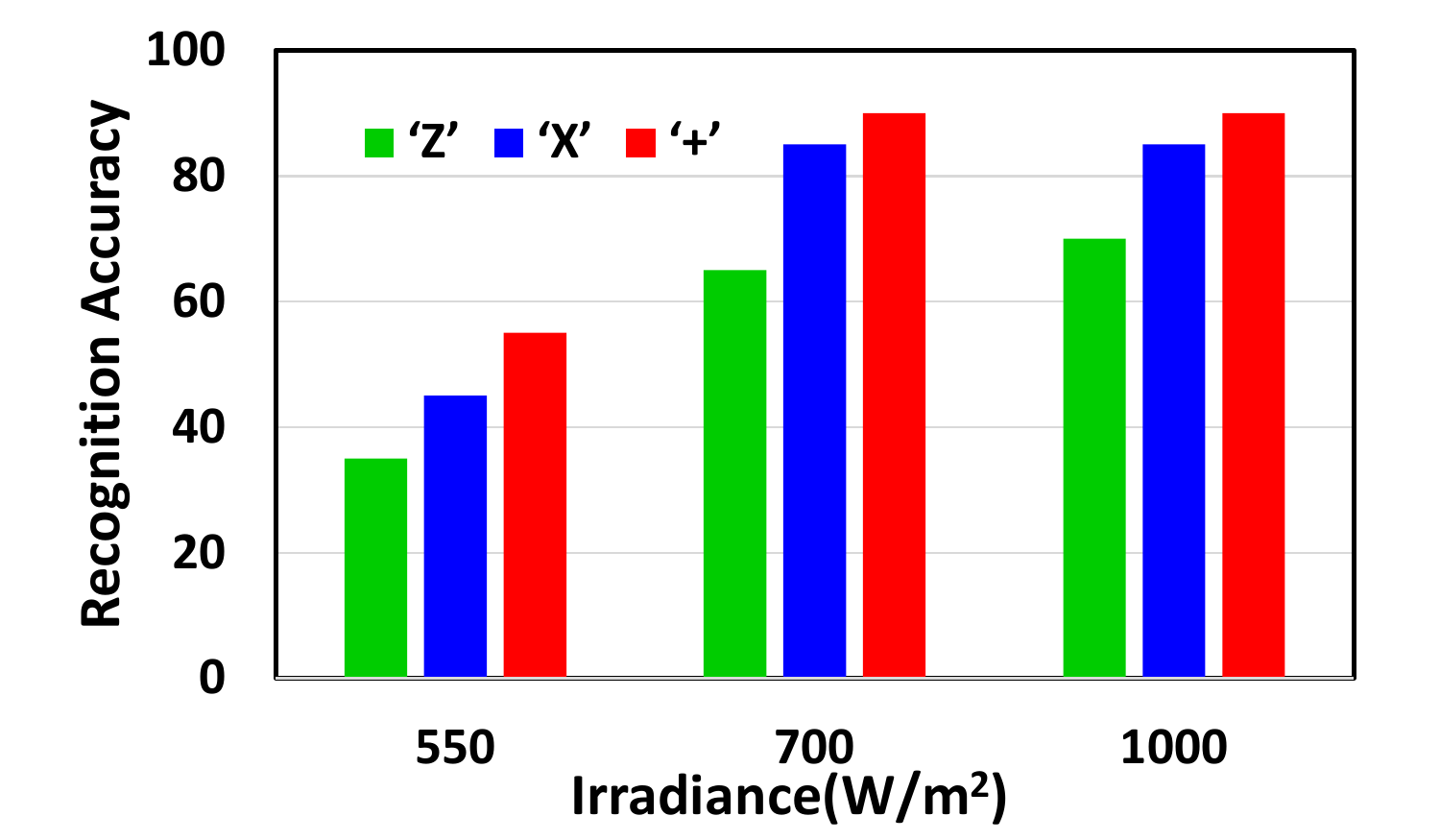}
\caption{Recognition accuracy vs Irradiance }
\label{recogacc}
\end{figure}

For different numbers of compressed measurements, we measure the energy consumption per frame and the recognition rate. We evaluate 20 gestures of each class, and the recognition rate is calculated from the total number of correctly recognized gestures. The total time per gesture is kept at 0.1secs. In a typical instance, the motion centers for a gesture ``Z'' as extracted from the hardware is shown in Figure \ref{Zt}. Comparison with Figure.~\ref{M_vs_MC} reveals a close match between simulation and measurement. Figure \ref{fig:power_nmeas} shows the measured design space exploration. We measure the recognition accuracy as a function of $M$ which reveals an accuracy rate of $>80\%$ for $M>300$, which closely matches simulation results described in the previous section. Figure \ref{fig:power_frame} shows dependence of the power consumed by the MCU and the corresponding recognition accuracy of the proposed system as a function of the frame rate. We note that a minimum frame rate of 5fps is required for maintaining a desired recognition accuracy of [$\geq$84\%]. As the frame rate increases, the corresponding power consumption also increases and shows a graceful trade-off between accuracy and power consumed. Figure~\ref{fig:Irradiance} illustrates the efficiency of the power management system where the irradiance of the incident light is varied. The corresponding power consumed and the maximum frame rate that can be supported is also shown. It can be noted that for an irradiance of $1000W/m^2$ (typical for outdoor sensors) a frame rate of 10fps and recognition accuracy of $>80\%$ is achieved. 

As the environmental conditions and irradiance levels change, the proposed system can scale the frame/sec accordingly, which gracefully trades-off recognition accuracy. Figure \ref{recogacc} illustrates the tradeoff between recognition accuracy for different gestures as a function of Irradiance.

\begin{table}[!h]
\begin{center}
 \begin{tabular}{|c| c| c| c|} 
 \hline
 Gesture Type & + & Z & X \\ 
 \hline
 Recognition Accuracy & 90\% & 70\% & 85\% \\ 
 \hline
\end{tabular}
\end{center}
\vspace{-0.5cm}
\caption{Recognition accuracy of 3 gesture classes (M=400)}
\label{recogaccuracy}
\vspace{-0.5cm}
\end{table}

\subsection{Multi-class Recognition Accuracy}

At $M = 400$ the recognition accuracies of 3 different gesture classes are shown in Table \ref{recogaccuracy}.
It can be seen that the recognition accuracy depends on  complexity of the gesture. For a simple gesture, e.g., ``+'', a peak accuracy of $90\%$ in a fully solar energy harvested system is measured. A comparison of the proposed system with competing hardware ~\cite{leecpu,chaofpga,yufpga,sohamwork} based motion and gesture detection is shown in Table. II.  The proposed system demonstrates more than $3\times$ improvement compared to reported works in energy/frame for detecting ``wake up'' gestures. This enables a fully self-powered ``always on'' camera front end.

\begin{figure}[!b]
\centering
\includegraphics[scale=0.32]{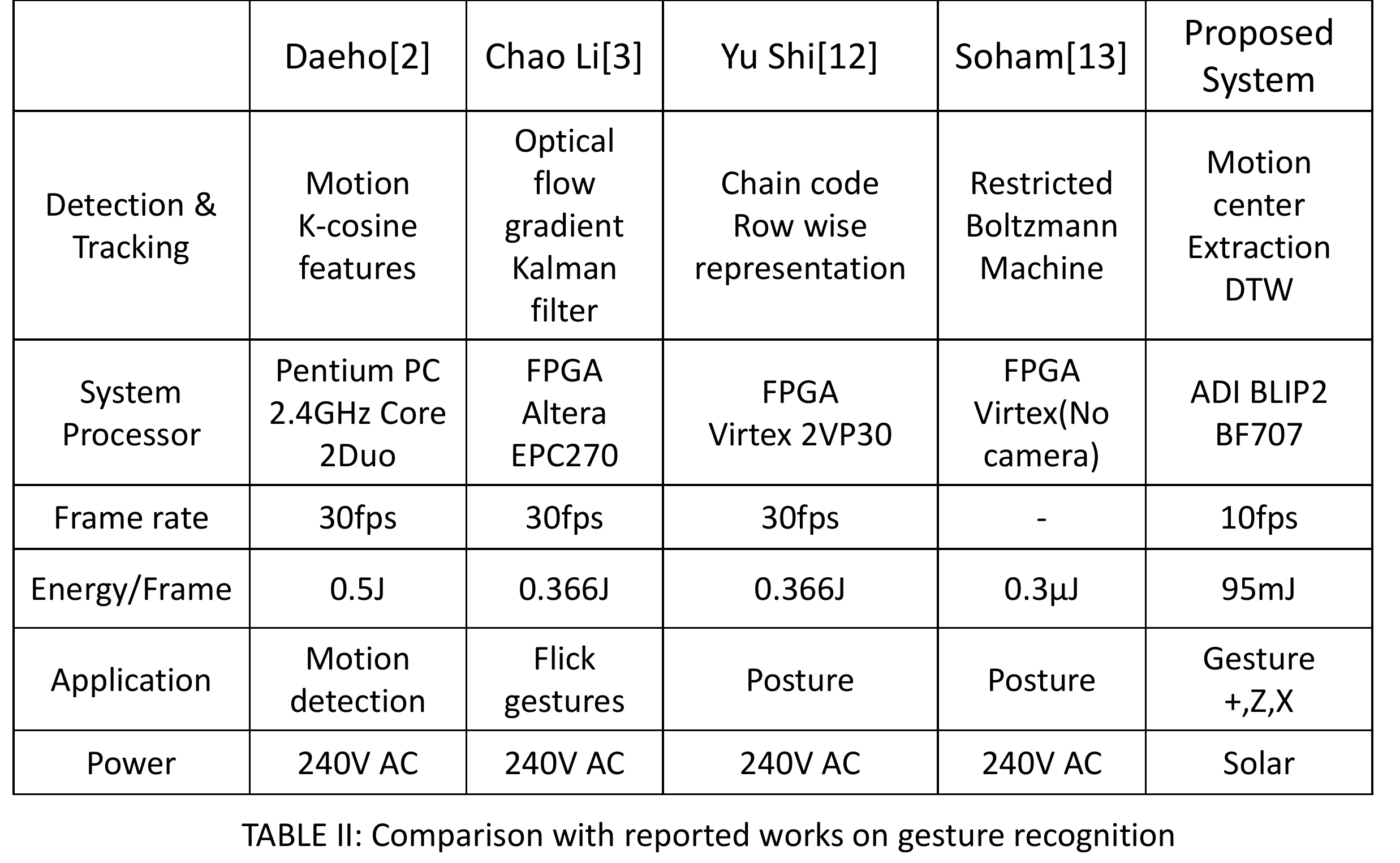}
\end{figure}


\section{Conclusions}
This paper presents a solar powered, ``always on'', gesture recognition system that provides a trigger for system ``wake up''. The major savings of power in our system comes from the two layers of compression that reduce the resolution of the image sensor by a factor of more than $768\times$ [$256\times$ by block averaging and $3\times$ by random compressive measurements]. The block compression layer preserves the geometric information of the gesture and the random projection layer preserves the motion parameters. These two preservations are the keys for maintaining a high recognition rate in the compressed domain. Further a hardware-algorithm co-design allows energy-efficient mapping of the recognition algorithm on a low power MCU and powered by a solar powered DC-DC converter and regulator with MPPT. The system demonstrates an average recognition accuracy of $>80\%$ while consuming less than $95mJ/frame$.

\section{Acknowledgment}
This work was funded in part by Intel Corp. and the NSF CRII Award 1464353.
\vspace{-0.2cm}


%

%
%

\end{document}